\documentclass[11pt]{article}
\usepackage[utf8]{inputenc}
\usepackage[T1]{fontenc}
\usepackage[english]{babel}
\usepackage{lmodern}
\usepackage{microtype}
\usepackage[hmargin=1in,vmargin=1in]{geometry}
\usepackage{amsmath}
\usepackage{enumitem}
\usepackage{graphicx}
\usepackage{longtable,booktabs,array}
\usepackage{calc}
\usepackage{float}
\providecommand{\real}[1]{#1}

\usepackage{xcolor}
\usepackage{hyperref}
\hypersetup{colorlinks=true, linkcolor=black, citecolor=black, urlcolor=[rgb]{0,0,0.6}}

\newcommand{\refentry}[1]{\noindent\hangindent=1.5em\hangafter=1 #1\par\vspace{0.45em}}

\title{Memory-Orchestrated Semantic System (MOSS): An Auditable Agentic Memory Architecture}
\author{Serge Lacasse\textsuperscript{*}, Jérémie Hatier\textsuperscript{\dag}, Alex Baker\textsuperscript{\dag}}
\date{}

\begin{document}
\maketitle

\begin{center}\small
\textsuperscript{*} Faculté de musique, Université Laval, Québec, Canada \\
\textsuperscript{\dag} Faculté des sciences et de génie, Université Laval, Québec, Canada \\
\emph{Correspondence: serge.lacasse@mus.ulaval.ca}
\end{center}

\begin{abstract}
Long-term memory remains a structural weakness of AI agents. The dominant approach, retrieval-augmented generation (RAG), relies on embedding-based similarity search, which is opaque by construction, difficult to audit, and bounded by the theoretical limits of vector representations. We present the Memory-Orchestrated Semantic System (MOSS), an agentic memory architecture in which the agent (not a similarity pipeline) drives retrieval over a structured relational database. MOSS is model-agnostic, storage-agnostic, and API-agnostic: it runs on any relational engine, connects to any LLM provider (or to deterministic non-LLM processes), and deploys on any infrastructure, local or cloud. Its retrieval execution is symbolic and reproducible---once a query is formulated, no LLM participates in the retrieval loop---and every step of the system, from indexing to answer formulation, is logged and inspectable, making MOSS auditable by construction. Rather than imposing an external ontology, MOSS derives its conceptual vocabulary inductively from the corpus itself. We report on a longitudinal deployment unique in the agentic-memory literature: roughly a year of continuous production (since May 2025) over the complete working corpus of an individual scholar---a conversational corpus reaching back to October 2024 (some 44 million tokens, retroactively indexed) comprising 110,183 segments, alongside 163,494 catalogued documents, 569 inductively derived concepts, 322,662 concept annotations, and eleven metadata graphs totalling approximately five million relations---across four successive infrastructure generations. While the present case is that of a single researcher, the architecture is in no way specific to one person: it serves a team, an institution, or any entity that accumulates knowledge over time. We argue that auditable, sovereign, structurally unbounded memory is a precondition for AI agents intended to accompany a person or an organization over years rather than sessions.
\end{abstract}

\noindent\textbf{Keywords:} agentic memory, long-term memory, AI agents, retrieval-augmented generation, auditability, data sovereignty, knowledge organization, exocortex

\section{Introduction}

Large language models are stateless. Whatever continuity an AI agent
exhibits across sessions must be supplied by infrastructure surrounding
the model; what recent literature calls the \emph{harness} [Zhang et
al., 2026]. The formula \emph{Agent = Model + Harness} has become
standard. Within that harness, memory is the component that determines
whether an agent can accompany, and interact with, a person or an
organization over months and years, or whether every conversation
begins, cognitively speaking, from zero. This paper centers on a single
deployed case---the working corpus of an individual scholar---but the
architecture is in no way specific to one person: the same machinery
serves a team, an institution, or any entity that accumulates a body of
knowledge over time (Section 5.3). The individual case is our starting
point, not our limit; it happens, as we argue, to be the hardest one.

The ambition itself is old. Bush [1945] imagined the memex, a device
holding all of a person's books, records, and
communications, navigable by association; Gemmell, Bell and colleagues
built MyLifeBits [Gemmell et al., 2002, 2006], storing one
researcher's entire digital life (notably in a SQL
database) two decades before large language models existed. What that
lineage lacked was not storage or structure but the \emph{navigator}: an
agent intelligent enough to traverse a lifetime of organized memory in
the service of ongoing thought.

The industry's default answer to agent memory did not
come from that lineage. It came from retrieval-augmented generation
(RAG): documents are chunked, embedded into vectors, and retrieved at
query time by similarity search [Gao et al., 2023]. RAG is powerful,
but it carries structural limitations---opacity, validation difficulty,
and formal bounds---that become acute precisely in the long-term,
personal, or regulated settings where memory matters most. We review
them in Section 2. Meanwhile, the research field of agent memory has
rapidly organized itself, with recent surveys explicitly delineating
agent memory from RAG and context engineering [Hu et al., 2025; Zhang
et al., 2024; Du, 2026] and empirical analyses finding that current
systems underperform their theoretical promise, on underscaled
benchmarks with misaligned metrics and ignored system costs [Jiang et
al., 2026]. The field thus both names the category this paper
inhabits---\emph{agentic memory} [Xu et al., 2025]---and calls for
the kind of evidence this paper supplies: sustained operational
experience rather than another benchmark entry.

The \textbf{Memory-Orchestrated Semantic System (MOSS)} is an agentic
memory architecture that has been growing through development and use
for about a year. Real production began on 9 May 2025; the system moved
to cloud infrastructure that summer and acquired its structured database
in the autumn of 2025. The conversational corpus it indexes, however,
reaches back to October 2024---those earlier exchanges, predating the
database, were incorporated retroactively. The agent analyzes the
intention behind a query, parameterizes a structured search, and
navigates a relational database that holds a richly indexed map of the
entire corpus---conversations and documents alike. The retrieval itself
is executed in SQL, deterministically and reproducibly, with no LLM in
the retrieval loop.\footnote{``SQL'' here denotes the relational query model, not a specific engine: MOSS is database-agnostic and runs on any relational backend (SQLite in the current deployment; PostgreSQL and distributed SQL engines equally).} Every call is logged in real time; the raw corpus
is preserved intact, human-readable, outside any proprietary format. The
fundamental design stance can be stated in one sentence: \textbf{the
database is the map; the documents are the territory.} MOSS retrieves
information through an explicit, inspectable map rather than
synthesizing it through an opaque process---and when the map suffices
(because segments carry summaries, concepts, participants, timestamps,
affective state, and more), which is most of the time, the territory
need not be visited at all.

Our contributions are threefold. \textbf{(1) An architecture} organized
around three commitments: \emph{auditability by construction} (every
retrieval is an inspectable SQL query, every system action is logged,
the source corpus remains human-readable); \emph{triple agnosticism}
(any relational engine, any LLM provider or non-LLM process, any storage
or execution infrastructure); and an \emph{inductive corpus ontology}
derived bottom-up from the corpus rather than imposed from outside.
\textbf{(2) A longitudinal production deployment}: to our knowledge, no
published agentic-memory system reports comparable longevity on a real
personal corpus---about a year of continuous production (since May
2025), four successive infrastructure generations, and a memory whose
conversational corpus reaches back to October 2024 (110,183 segments,
$\approx$44M tokens, retroactively indexed) alongside 163,494 catalogued
multimodal documents, interrogated daily as the primary working memory
of an active scholar. \textbf{(3) A positioning} of the state of the art
as a sequence of progressively closer neighborhoods---from classical
vector RAG, through graph-augmented retrieval and production memory
layers, to the recent convergence of agentic memory with structured,
SQL-native, locally owned storage---showing that MOSS belongs to this
last and most recent group while differing from every member on at least
one structural axis, and occupies an intersection (a lifetime corpus, an
agentic navigator, an auditable relational substrate) that no published
system covers.

Section 2 reviews the state of the art, from the most distant paradigm
to MOSS's own group. Section 3 describes the
architecture. Section 4 reports the longitudinal deployment. Section 5
discusses implications: auditability, sovereignty, and the individual
knowledge worker as the ontologically hardest case of memory
engineering. Section 6 presents the \emph{métacalque} overlay layer
under development. Section 7 outlines future work.

\section{From Retrieval Pipelines to Agentic Memory: Situating MOSS}

We organize related work as four concentric circles, from the paradigm
most distant from MOSS to the group MOSS belongs to. At each circle we
state what the group contributes and where MOSS departs from it.

\subsection{Farthest: classical vector RAG is not a memory}

The RAG paradigm---chunk, embed, retrieve by cosine similarity, inject
[Gao et al., 2023]---is a \emph{retrieval pipeline}, not a memory
system: it has no organization of experience, no temporality, no
identity of the corpus, no account of how knowledge accumulates. This is
no longer a contrarian claim; the field's own surveys
now delineate agent memory from RAG [Hu et al., 2025]. Three
structural limitations are documented. RAG is \textbf{opaque}: cosine
proximity in a high-dimensional space explains nothing a human can
verify. It is \textbf{unvalidatable by design}: studies of real
deployments conclude that RAG systems can only be validated in
operation, robustness emerging from iteration rather than design
[Barnett et al., 2024]. And it is \textbf{theoretically bounded}:
single-vector retrieval has formal limits no engineering can remove
[Weller et al., 2025], with follow-up work showing that embedding
similarity behaves as a noisy statistical proxy for relevance whose
failures worsen as the corpus grows---relevant documents are
progressively ``drowned'' [Archish et al., 2026]---and that
performance scales unpredictably with embedding dimension outside the
training distribution [Killingback et al., 2026]. A practical
concern compounds these: the vector index is captive to the embedding
model that produced it, a representational lock-in. On very long-term
conversational benchmarks, RAG with long context windows remains far
from human performance on temporal and causal reasoning [Maharana et
al., 2024]; in high-stakes domains, probabilistic retrieval has been
argued operationally unacceptable outright [Agand, 2026].

\emph{Departure.} MOSS is not a refinement of this pipeline; it is a
memory system in the cognitive-science sense---encoding (both real-time
and retroactive indexing), organized storage (a relational map with an
ontology), and agent-driven recall---in which embedding similarity plays
no role in the retrieval path. Where RAG performs \emph{document}
retrieval, MOSS retrieves \emph{information}.

\subsection{Closer: graph-augmented retrieval (GraphRAG)}

A first move beyond flat similarity is to structure knowledge as
entity--relation graphs. GraphRAG is now a mature subfield with major
surveys [Peng et al., 2024; Zhang et al., 2025; Zhu et al., 2025], a
strong temporal current---timestamped relations and validity intervals
for evolving knowledge [Rasmussen et al., 2025; Han et al., 2025; Yang
et al., 2026]---and agentic variants in which a model explores the
graph during retrieval [Sanmartin, 2024; Luo et al., 2025].
Graph-based agent memory has its own survey [Yang et al., 2026].
This group understood two things MOSS shares: relations matter, and time
matters.

\emph{Departure---threefold.} First, the \textbf{unit of relation}:
GraphRAG \emph{distills} documents into extracted entity--relation
triples; the map replaces the territory, lossily. MOSS's
eleven graphs (\textasciitilde5 million typed relations) connect
\emph{indexed database segments}---never raw extracts---and replace
nothing: they are a navigation layer over the map, pointing into a
territory kept intact. Second, the \textbf{traversal}: agentic GraphRAG
places the LLM inside the retrieval loop (exploring nodes step by step),
sacrificing reproducibility; MOSS traverses relations with recursive
SQL, deterministically.\footnote{In the current SQLite deployment, graphs are stored as relation tables and traversed with recursive SQL, requiring no dedicated graph engine. On engines with native graph support (e.g., PostgreSQL with its graph extensions), the same logical traversal can be delegated to the engine; the architectural commitment---deterministic traversal with no LLM in the loop---is unchanged.} Third, the \textbf{relation families}:
alongside co-occurrence, native temporal adjacency (every segment is
timestamped at the source, not retro-dated), and thematic cohesion, MOSS
maintains \emph{affective resonance} edges---a relation type absent from
this literature.

\subsection{Closer still: production memory layers}

A third group treats memory as a product in its own right.
\textbf{Letta/MemGPT} [Packer et al., 2023] pioneered the
operating-system metaphor, the LLM paging its own memory between context
and external storage. \textbf{Mem0} [Chhikara et al., 2025], the
market reference, combines dynamic fact extraction, a knowledge graph,
and vector retrieval. \textbf{Zep} [Rasmussen et al., 2025] builds a
temporal knowledge graph with explicit fact-validity intervals,
outperforming MemGPT on deep memory retrieval. Knowledge-graph personal
memory frameworks extend this line [Menschikov et al., 2026]. And
platform vendors now ship managed memory: OpenAI's
\emph{Dreaming} (V3, June 2026) continuously synthesizes a user-context
profile from conversation history in a background process, injected into
every new session. This group understood that memory is the
differentiating layer of the agent stack, and its drift---toward
structure and temporality---is unmistakable.

\emph{Departure.} These systems remain vectorial or hybrid-vectorial in
the retrieval path (Mem0), proprietary in representation
(Zep's graph engine; Dreaming's hosted
profile), or model-managed (MemGPT's self-paging). None
claims end-to-end auditability; none separates \emph{formulation}
(agentic) from \emph{execution} (symbolic). The sharpest contrast is
with background synthesis: \textbf{Dreaming synthesizes an opaque
profile the user can neither inspect nor take elsewhere; MOSS retrieves
verifiable information from an inspectable map the user owns at every
layer.} Existing systems either retrieve documents by similarity or
synthesize memories opaquely; MOSS retrieves
\emph{information}---deterministically in execution, auditably
end-to-end, sovereignly.

\subsection{MOSS's group: agentic memory meets the structured, local turn}

The most recent literature converges from two directions onto the ground
MOSS occupies. From one side, \textbf{agentic memory} is now a named
category: A-MEM [Xu et al., 2025] organizes memories as an evolving
Zettelkasten-style network under agent control; AgeMem [Yu et al.,
2026] exposes memory operations as tool actions within the
agent's own policy; and ``beyond RAG for agent memory'' is
now a stated thesis, that flat similarity retrieval is structurally
mismatched to an agent's bounded, coherent interaction
stream [Hu et al., 2026]. From the other side, a \textbf{structured,
SQL-native, locally owned turn} is underway in systems engineering:
Memori (GibsonAI, 2025) stores extracted facts and preferences in
standard SQL databases, marketing transparency and user ownership;
SuperLocalMemory [Bhardwaj, 2026] is local-first SQLite memory with
full-text search and no LLM calls at retrieval time; AnnoRetrieve [Lin
et al., 2026] replaces embeddings altogether with structured
annotations over automatically induced schemas, queried in SQL; AgentSM
[Biswal et al., 2026] gives text-to-SQL agents an interpretable
structured memory of execution traces.

MOSS belongs to this group---it is an agentic memory with a SQL-native,
locally ownable substrate---and predates much of it in production.
Within the group, five differences remain:

\begin{enumerate}
\def\labelenumi{\arabic{enumi}.}
\item
  \textbf{Scale and nature of the corpus.} Memori and SuperLocalMemory
  manage session-scale facts and preferences; AnnoRetrieve indexes
  enterprise document collections. MOSS indexes the \emph{complete
  working corpus of a person or organization}---every conversation turn
  and every document of an intellectual life---under a single
  organization.
\item
  \textbf{Inductive ontology.} MOSS's 569 concepts are
  derived bottom-up from the corpus and applied systematically (322,662
  annotations), in the spirit of codebook thematic analysis [Braun \&
  Clarke, 2006]---neither an imposed schema nor mere entity extraction
  (Section 3.3).
\item
  \textbf{Query sovereignty.} Relevance weighting is a property of the
  \emph{question}, decided by the agent per query (Section 3.4), not
  baked into the index at storage time. No system in the group performs
  this inversion.
\item
  \textbf{Native affective and temporal indexing.} Valence and
  activation are first-class attributes of every conversational segment,
  alongside source timestamps.
\item
  \textbf{Longitudinal evidence.} About a year of continuous production
  (over a corpus reaching back further still) across four infrastructure
  generations---precisely the operational evidence the
  field's own empirical critique calls for [Jiang et
  al., 2026], and that no system in the group reports.
\end{enumerate}

\subsection{A note on the word ``semantic''}

Industry usage calls embedding-based similarity ``semantic search,'' to
distinguish it from earlier ``lexical'' approaches [Huyen, 2024, pp.
258--261]. Strictly, what embeddings implement is
\emph{distributional} semantics---meaning as statistical co-occurrence
[Sahlgren, 2008; Lenci, 2018]---and the limits of equating form with
meaning are well argued [Bender \& Koller, 2020]; even the
vector-database literature lists ``the ambiguity of semantic similarity''
as a central obstacle [Pan et al., 2023]. The word \emph{semantic}
entered artificial intelligence with the opposite referent: explicit,
structured, relational meaning---semantic networks [Quillian, 1968],
semantic memory [Tulving, 1972]. The \emph{S} in
MOSS---\emph{Semantic}---is meant in that original sense: explicit
concepts, typed relations, symbolically queryable structure. We
therefore distinguish throughout between \emph{distributional}
(geometric, probabilistic) and \emph{structural} (relational, symbolic)
semantics; MOSS implements the latter. This is not only a conceptual
preference: on our own corpus, classical distributional methods were
implemented and then set aside in favor of structured extraction, on the
evidence of output quality (Section 3.3, note 3).

\subsection{The vacant intersection}

Three research traditions each hold one piece of what a lifetime agent
memory requires. The lifelogging lineage---memex, MyLifeBits, and
successors [Bush, 1945; Gemmell et al., 2002; 2006; Gurrin et al.,
2014]---captured the \emph{corpus of a life}, even in SQL, but
predates intelligent navigators. The agent-memory systems of Sections
2.3--2.4 have the \emph{agent} but operate on session-scale stores. The
structured turn has the \emph{auditable substrate}, but not the lifetime
corpus. MOSS occupies the intersection of the three: a lifetime corpus,
an agentic navigator, an auditable relational substrate. It can be read
as the agentic heir of MyLifeBits---the database of a life, finally
given the navigator the memex always lacked.

\section{Architecture}

\subsection{Design principles}

Three commitments govern every design decision in MOSS.

\textbf{The database is the map; the documents are the territory.} The
raw corpus (conversation transcripts and documents) is never altered,
chunked into oblivion, or replaced by derived representations. It is
preserved intact in ordinary files on ordinary storage, readable by any
human at any time. Above it, a relational database holds a dense map:
segments with summaries, timestamps, participants, affective state, and
concept annotations---or any other field of information relevant to a
given application; documents with summaries, keywords, and structural
outlines; graphs of relations. Most queries are answered from the map
alone, at a fraction of the token and energy cost of re-reading sources;
the territory is consulted only when genuinely necessary.

\textbf{The query is sovereign.} Relevance is not a static property of
stored items; it is a property of the question being asked. The same
segment may matter enormously for a temporal question, marginally for a
thematic one, and not at all for an affective one. Accordingly,
weighting in MOSS is decided at query time, per query, by the
agent---not baked into the index at storage time. This inverts the RAG
assumption that a single embedding-space geometry can serve all
questions.

\textbf{Audit everything; hide nothing.} Every call into the system is
logged in real time. Every retrieval is an explicit SQL query that can
be re-executed and inspected. Conversation archives are plain text with
timestamps and token counts. The answer to ``why did the agent say this?''
is always reconstructible from logs and queries---there is no layer at
which the explanation dissolves into geometry.

These commitments are instances of a broader epistemic framework of
memory---axioms, properties, and operations that a memory system must
satisfy in order to reason with itself---which we develop in a companion
paper.

\subsection{System overview}

MOSS consists of two components (Figure 1).

The \textbf{Orchestrator} contains all active code: a REST API and an
MCP (Model Context Protocol) endpoint; connectors to LLM providers; the
QueryProfiler (Section 3.4); and the SQL retrieval layer. It is the
single connection point between agents and memory.

The \textbf{Memory Structure} is the relational database itself,
together with the conventions that organize it.\footnote{By \emph{conventions} we mean the instruction layer that governs the agents' behavior---what we elsewhere call the system's \emph{constitution}: system prompts, operating playbooks, and tool-use protocols. These are versioned configuration, distinct from the data they operate on.} It runs on any
relational engine and contains the segment metadata, the document
catalogue, the inductive concept ontology, and the relation graphs.

Both components are agnostic on three axes. \emph{Model-agnostic}: the
Orchestrator connects to hosted providers (Google AI Studio and Vertex
AI, Azure OpenAI, the Anthropic API) and to local models (Ollama, LM
Studio); the same memory has served GPT-4o, Gemini, and Claude
interchangeably, and nothing requires the consumer to be an LLM at
all---a deterministic process (for example, a sensor pipeline writing
observations) can be a memory client. \emph{Storage-agnostic}: the raw
corpus may live on consumer cloud storage, enterprise object storage
(Azure Blob, S3), Microsoft 365, or on-premise servers.
\emph{API-agnostic and bidirectional}: MOSS can act as the orchestrator
that calls a model through its API, or be mounted as a resource inside
an existing agent environment through MCP---in the latter configuration
the dependency is inverted, and a commercial assistant (e.g., Claude)
connects \emph{into} the user's sovereign memory.
\begin{figure}[tbp]
  \centering
  \includegraphics[width=\linewidth]{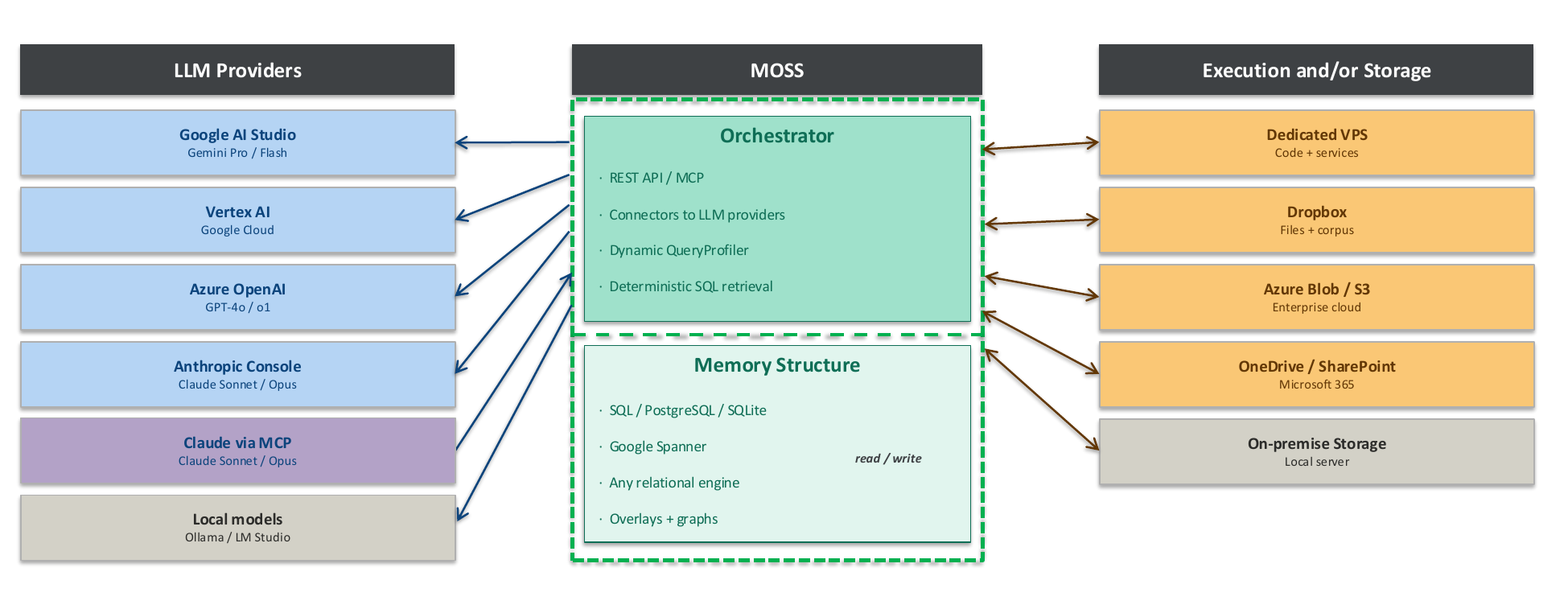}
  \caption{MOSS deployment architecture. The Orchestrator holds all active
  code---REST API / MCP, connectors to LLM providers, the dynamic
  QueryProfiler, and deterministic SQL retrieval---and is the single point of
  contact between agents and memory. The Memory Structure is the relational
  database itself. The system is model-agnostic (left: hosted and local
  providers), engine-agnostic (any relational database, centre), and
  storage-agnostic (right: cloud, enterprise object storage, Microsoft~365,
  on-premise). Deployment is local or cloud, with no vendor lock-in, and
  compatible with any existing infrastructure.}
  \label{fig:deploy}
\end{figure}

\subsection{The structured memory}

\textbf{Segments.} Conversations are segmented semantically, in real
time for live exchanges and retroactively from imported archives (JSON
exports from three assistant platforms, dating back to October 2024).
Each of the 110,183 segments carries a precise timestamp, a reference
token count, a generated summary, the persons mentioned, an affective
state encoded as valence and activation, and its concept annotations.
Segments are the primary retrieval unit; their metadata makes most
questions answerable without touching raw text.

\textbf{Documents.} 163,494 files are catalogued (documents, code,
audio, video, images). Of these, approximately 21,000 text documents
(PDF and DOCX prioritized) are enriched with summaries, keywords,
structural outlines, and concept annotations.\footnote{Image-only PDFs are currently being transcribed to text (via Docling) for indexing, which is expected to bring the enriched-text count to roughly 40,000 documents; multimodal enrichment of audio, video, and image holdings is in development. The system is under continuous extension.} Document outlines enable
hierarchical navigation: the agent reads the outline, reasons about
which sections are likely to carry the answer, and reads only those.
Notably, each node of the outline carries its own short summary, so the
agent reasons not over bare section titles but over a glossed,
hierarchical map of the document's contents. The
effectiveness of this reasoning-over-structure pattern against dense
vector retrieval has been independently demonstrated in recent
vectorless-RAG work.

\textbf{The inductive ontology.} MOSS does not impose an external
controlled vocabulary.\footnote{The extraction pipeline was selected empirically. We first implemented classical distributional approaches (Word2Vec, then FastText); on our corpus their concept outputs proved weaker than direct LLM keyword extraction, which we retained, with folder-level grouping. Distributional embedding was thus not bypassed on principle but evaluated and set aside on the evidence---bearing on the distinction in Section 2.5. External controlled vocabularies were likewise tested and abandoned: schemes such as the Library of Congress Subject Headings, the FAST faceted vocabulary, and Roget-style thesauri never matched the precision with which an inductively emerged ontology fits a memorial corpus of this kind.} Its 569 concepts have three provenances: a small
seed of foundational concepts hand-injected at the outset, an
automatically extracted layer that constitutes the large majority, and a
set of named entities (organizations, institutions, tools) added
separately. The extracted layer was derived inductively from the corpus
itself and then applied systematically---322,662 segment-level
annotations in total. This mirrors, deliberately, the logic of codebook
thematic analysis in qualitative research [Braun \& Clarke, 2006]:
codes emerge from the data (inductive property) and are then applied
consistently across the corpus (codebook property). The hand-injected
seed---a handful of domain-specific concepts that form the conceptual
skeleton from which the extracted ontology grew---is an acknowledged
human starting point rather than a pretense of pure automation. The
result is an ontology that fits its corpus exactly (including
idiosyncratic concepts no external taxonomy would contain) while
remaining queryable as ordinary relational data. Themes, in this design,
are not pre-computed: they are assembled by the agent at query time by
grouping concepts relative to the question, consistent with the
principle that the query is sovereign.

\textbf{Graphs.} Eleven metadata graphs connect database segments (never
raw files) with approximately five million typed relations:
co-occurrence, temporal adjacency, affective resonance, thematic
cohesion, and geographic anchoring (in development). Where GraphRAG
distills documents into extracted triples that stand in for the sources,
MOSS's graphs relate the \emph{indexed segments
themselves} and point back into an intact territory. The graphs are
produced by two complementary mechanisms: a comprehensive weaving
performed at each system restart, and on-demand weaving triggered when a
specific query benefits from a freshly computed relational view. To our
knowledge, the affective-resonance family (edges connecting segments
with similar valence--activation signatures) has no equivalent in the
graph-memory literature.

\textbf{Overlays and the \emph{métacalque}.} Several of the objects
described above---the segmentation into conversational units, the
concept annotations, the affective state---are not a base structure with
decorations added on top. They are \emph{calques} (overlays):
transparent strata, each tracing one angle of reading directly over the
corpus. What unifies them is not their sum but the \emph{métacalque}
(which could be rendered as ``meta-overlay''): the principle that any
point of the corpus may be re-read through arbitrary, superimposable
overlays, none of them foundational.\footnote{\emph{Métacalque}, from French \emph{calque} (``tracing, overlay''): a semantic stratum traced over the corpus, in the cartographic sense of a transparent sheet laid over a map---through which the territory remains visible, and which can be removed or combined with others, no one of them beneath the rest. We retain the French-derived term as a term of art, since English ``layer'' collides with unrelated senses (network layers, meta-learning) and connotes a vertical stack rather than a superimposed, removable, combinable tracing. A fuller, operational definition is given in Section 6.} Segmentation, on this view, is
one overlay among others---the first the system happened to build, not
the substrate the others rest upon. (The relation graphs are not
overlays: rather than tracing a stratum over the text, they weave links
between metadata, a complementary mechanism.) This unification matters
for two reasons. First, it is economical: rather than an accumulating
inventory of heterogeneous structures, MOSS exposes one kind of object
that takes many forms, so that adding a capability (an overlay that
flags quotations or code blocks, an affective overlay over deposition
transcripts, a topical overlay) is the same architectural act each time,
not a new subsystem. Second, it is consistent with query sovereignty:
each overlay answers a distinct class of question. The segment overlay
is far from incidental---it answers questions of conversational
localization (``in which of my exchanges does this element appear?''), a
frequent and useful class; the concept overlay answers thematic
questions; the affective overlay, emotional ones. No overlay is
fundamental; each is the right instrument for a class of queries. The
structures described in this section are those serving the production
system today; a unified overlay layer that re-expresses and extends them
is under active development and is described in Section 6.

\subsection{The query lifecycle}

A query traverses MOSS as follows (Figure 2):

\begin{enumerate}
\def\labelenumi{\arabic{enumi}.}
\item
  \textbf{Intent analysis (agentic).} The agent receives the question
  and analyzes its intention: temporal (``when did we first\ldots''),
  thematic, affective (``when was I most worried about\ldots''), personal,
  documentary, or a combination.
\item
  \textbf{Profiling (agentic $\rightarrow$ structured).} The agent issues a
  natural-language request to the \textbf{QueryProfiler}, with a
  structured weighting appropriate to the intent: timestamps for
  historical questions, valence/activation for affective ones, concepts
  for thematic ones, or any combination. The QueryProfiler translates
  this profile into SQL. For needs beyond the profiler's
  vocabulary, the agent may write SQL directly (accepting occasional
  syntax retries---the very cost the QueryProfiler exists to eliminate
  in the common case).
\item
  \textbf{Execution (deterministic).} The SQL retrieval layer executes
  the query. \textbf{No LLM participates in this step.} Given the same
  query and database state, the result is identical, reproducible, and
  explainable line by line.
\item
  \textbf{Evaluation and iteration (agentic).} The agent inspects the
  results and, if unsatisfied, reformulates---changing weights,
  combining parameters, crossing dimensions, or revising the strategy
  entirely.\footnote{The current deployment caps reformulation at fifteen iterations; this bound is a tuning parameter chosen as appropriate for the present system, not an architectural constant, and may be set differently for other contexts.} Once satisfied, it formulates the answer, citing what it
  found.
\end{enumerate}

The division of labor is the architectural point: \emph{formulation} is
agentic and benefits from the model's intelligence;
\emph{execution} is symbolic and inherits the determinism, auditability,
and energy efficiency of a database engine. The non-determinism of the
LLM is confined to the edges of the loop, where it belongs, and every
decision leaves an inspectable trace (the profile requested, the SQL
that ran).

\begin{figure}[H]
  \centering
  \includegraphics[width=\linewidth]{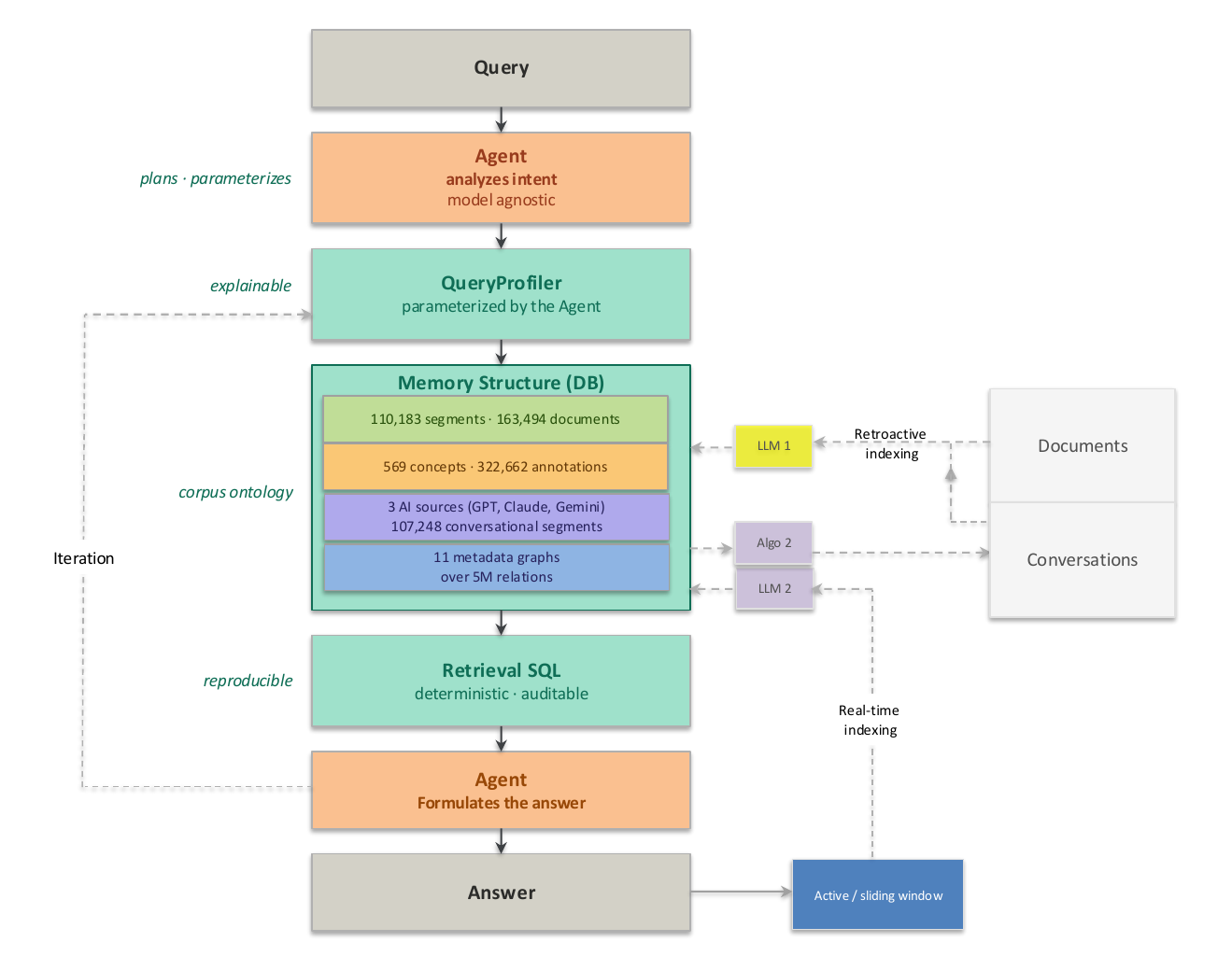}
  \caption{The query lifecycle and continuous indexing. Reading (centre): the
  agent analyzes the query's intent and parameterizes the QueryProfiler, which
  resolves the request through deterministic, auditable SQL retrieval over the
  Memory Structure, with no LLM in the loop; the agent then evaluates the
  results, reformulating if needed (dashed iteration loop) before formulating
  the answer. Non-determinism is confined to the agentic edges, while the
  deterministic core is traversed but never iterated. Writing (right): content
  leaving the active (sliding) window is indexed into the database in real time
  and archived to the conversation file, while imported documents and
  conversation histories are indexed retroactively---so the memory grows
  continuously and remains structurally unbounded.}
  \label{fig:lifecycle}
\end{figure}

\subsection{Real-time indexing and archiving}

As live conversation flows through the agent's working context ($\approx$35K tokens), that context is a finite, \textbf{sliding window}---a rolling buffer of recent turns, where each new exchange pushes the oldest out the far end. In a conventional setup, displaced content is lost: the window is a hard ceiling on what the agent can recall. MOSS removes that ceiling. A dedicated scribe agent intercepts everything leaving the active window and performs two operations on it: it \textbf{indexes} it into the database (segmentation, summarization, affective and conceptual annotation) and it \textbf{archives} it by appending to the conversation file for that day.\footnote{The day is used as the archival grouping unit here because it suits a scholar's daily working rhythm; another deployment could group segments by case, project, session, or any other unit appropriate to its domain.} Imported histories follow the same path retroactively through batch parsers.

Because nothing that scrolls out is lost---within seconds each fragment is a retrievable row in the structured memory---the finite window stops bounding what the agent can recall: anything older than the current context is one query away. From the agent's standpoint, the context is \emph{effectively unbounded}; it converses as though its memory were continuous and limitless, while the real window stays small, cheap, and fast. This is the operational meaning of the \emph{structurally unbounded} memory claimed in the abstract---not a bigger window, but a window whose contents never disappear. Keeping the window small carries a second benefit: the agent never accumulates a context long enough to suffer the \emph{lost-in-the-middle} degradation of very long windows [Liu et al., 2024], where information buried mid-context is poorly attended to---MOSS retrieves a handful of relevant rows rather than flooding the window. The result is a memory that grows continuously, with no gap between what was said and what is findable, and an archive that remains readable without any software at all.

\subsection{Auditability by construction}

Auditability in MOSS is not a logging feature bolted onto an opaque
core; it is the sum of properties already described: (i) retrieval is
explicit SQL, re-executable at will; (ii) every system call is logged in
real time, so the agent's full behavior---what it asked,
what it received, how it iterated---is observable as it happens; (iii)
the corpus of record is plain text with timestamps and token counts;
(iv) the ontology is a table, not a latent space. A compliance officer,
a researcher, or the user herself can answer, for any response the
system ever produced: \emph{what was retrieved, by which query, weighted
how, and why}. We know of no embedding-based or synthesis-based memory
system for which that sentence is true.

\section{A Year in Production: A Longitudinal Deployment}

MOSS has not yet been evaluated on a benchmark corpus; it has been
\emph{lived in}.\footnote{\label{fn:bench}While no single benchmark fits a system of this kind as a whole, its components map onto appropriate existing benchmark types, which we plan to use as evaluation proceeds: document-QA settings used for reasoning-based, vectorless retrieval (e.g., FinanceBench-style or multi-hop document QA) for its reasoning over structure; forgetting-aware and knowledge-update tasks (e.g., Memora/FAMA, LongMemEval) for its handling of superseded facts; and organization-oriented benchmarks (e.g., StructMemEval) for its claim that memory must be structured, not merely stored. Evaluation is matched capacity by capacity; only the integrated system awaits the comparative benchmarks its category will eventually invite.} Production began in May 2025; since
then the system has served the complete working memory of an active
scholar---research, teaching, administration, grant writing, technical
development, and personal organization---interrogated daily as primary
infrastructure, over a conversational corpus that reaches back to
October 2024 and was incorporated retroactively. We report this
deployment as an engineering case study [cf. Barnett et al., 2024]:
the strongest claims about memory systems are the ones a system survives
in operation---and this is the form of evidence the
field's own empirical critique identifies as missing
[Jiang et al., 2026].

\subsection{Four generations}

\begin{longtable}[]{@{}
  >{\raggedright\arraybackslash}p{(\columnwidth - 8\tabcolsep) * \real{0.1239}}
  >{\raggedright\arraybackslash}p{(\columnwidth - 8\tabcolsep) * \real{0.1022}}
  >{\raggedright\arraybackslash}p{(\columnwidth - 8\tabcolsep) * \real{0.1977}}
  >{\raggedright\arraybackslash}p{(\columnwidth - 8\tabcolsep) * \real{0.3813}}
  >{\raggedright\arraybackslash}p{(\columnwidth - 8\tabcolsep) * \real{0.1949}}@{}}
\toprule\noalign{}
\begin{minipage}[b]{\linewidth}\raggedright
\textbf{Generation}
\end{minipage} & \begin{minipage}[b]{\linewidth}\raggedright
\textbf{Period}
\end{minipage} & \begin{minipage}[b]{\linewidth}\raggedright
\textbf{Model layer}
\end{minipage} & \begin{minipage}[b]{\linewidth}\raggedright
\textbf{Orchestrator}
\end{minipage} & \begin{minipage}[b]{\linewidth}\raggedright
\textbf{Storage}
\end{minipage} \\
\midrule\noalign{}
\endhead
\bottomrule\noalign{}
\endlastfoot
\textbf{V0} & May 2025 & Custom GPT (ChatGPT) & Flask on shared Python
hosting; file access via cloud-storage API; no SQL, no MCP & Consumer
cloud storage, raw text files \\
\textbf{V1} & Summer 2025 & GPT-4o (Azure OpenAI) & Migration to Azure
Data Lake Gen2; atomic append; hierarchical structure & Azure Data Lake
Gen2 \\
\textbf{V2} & Autumn 2025 & Gemini (Google AI Studio) & Local
workstation (Apple Silicon); QueryProfiler; deterministic SQL retrieval;
metadata database created & Cloud storage + Data Lake; SQLite
(\textasciitilde4 GB) \\
\textbf{V3} & 2025--2026 & Gemini + Claude (via MCP) + local models &
Dedicated VPS; REST API + MCP; three services running 24/7 & Cloud
storage (168K files); SQLite (\textasciitilde5 GB) \\
\end{longtable}

The same architectural commitments---map over territory, query
sovereignty, full logging---survived all four migrations, across three
LLM vendors, two cloud providers, a local workstation, and a dedicated
server. We take this as empirical evidence for the agnosticism claims of
Section 3.2: the memory outlived every component around it.

\subsection{The memory at the time of writing}

As of June 2026, the production database contains:

\begin{itemize}
\item
  \textbf{110,183 conversational segments}, semantically segmented,
  individually summarized, affectively scored, and concept-annotated,
  drawn from \textasciitilde44 million tokens across roughly 600 files
  and three assistant platforms (ChatGPT, Claude, Gemini), spanning
  conversations from October 2024 to the present;
\item
  \textbf{163,494 catalogued files}, of which \textasciitilde21,000 text
  documents are fully enriched (summary, keywords, outline, concepts);
\item
  \textbf{569 inductively derived concepts} with \textbf{322,662 segment
  annotations};
\item
  \textbf{11 metadata graphs} totalling \textbf{\textasciitilde5 million
  typed relations};
\item
  a complete plain-text conversation archive with timestamps and token
  counts.
\end{itemize}

\subsection{Operational observations}

Three observations from sustained use are worth recording, as they are
invisible to benchmark evaluation.

\textbf{The map usually suffices.} Because segments carry summaries and
rich metadata, a large fraction of everyday questions---``when did we
first get MOSS running?'', ``synthesize my work on X since 1995'', ``when
was I most enthusiastic about this project?''---are answered entirely
from the database, in a handful of targeted rows, without opening a
single source file. The token economy relative to chunk-stuffing RAG is
structural, not incremental: the system retrieves two or three
\emph{right} segments instead of fifty chunks ``just in case.''

\textbf{Affective indexing earns its place.} Encoding valence and
activation per segment---an unusual choice in engineering-driven
systems---has proven one of the most used retrieval dimensions in
practice: questions about one's own history are very
often affective (``when was I worried, and about what?''), and a memory of
a human life that cannot answer them is incomplete.\footnote{Affective state is encoded on Russell's circumplex model of affect [Russell, 1980] (valence $\times$ activation), refined following Mattek et al. [2017], whose model has the legitimate range of valence narrow as activation falls---affect occupies a triangular region of the plane rather than a disc---so that low-arousal states are treated as less semantically determinate, and named less precisely, than high-arousal ones. Though this overlay may appear to bear only on narrowly personal matters---a subject's everyday mood, say---the same valence $\times$ activation substrate makes it, perhaps surprisingly, a domain-general instrument for the affective coloration of any verbal account: the demeanor legible in verbatim courtroom testimony, the affective shading with which patients (or their clinicians) describe a condition in clinical and psychiatric notes, or the appraisive register of evaluative responses in the arts, such as a listener's free-text assessment of a piece of music or a work of visual art.}

\textbf{Auditability changes the relationship to the system.} Because
every retrieval can be watched in the live log and every answer traced
to queries and segments, failures are diagnosable---a wrong answer
points to a wrong query or a gap in indexing, both fixable---rather than
mysterious. A year of daily trust in the system rests less on its
accuracy than on its \emph{inspectability when inaccurate}.

\subsection{Limitations of the evidence}

We state plainly what this deployment does and does not establish. It is
a single-corpus, single-user, longitudinal case study: it demonstrates
feasibility, durability, architectural portability, and sustained
practical utility at realistic personal scale. It does not provide
controlled comparisons against Mem0, Zep, or LoCoMo-style benchmarks;
those evaluations are planned (note 9 above, Section 7 below). The
deployment's irreproducibility is also its evidentiary
value---a year of one scholar's actual cognitive life,
over a corpus reaching back further still, cannot be re-run under
laboratory conditions---and we present it as the complement to, not a
substitute for, benchmark science.

\section{Discussion}

\subsection{Retrieval versus synthesis}

The June 2026 arrival of background memory synthesis at platform scale
(OpenAI's Dreaming~3) sharpens the
field's central fork. Synthesis-based memory is
convenient and increasingly capable, but it is \emph{constitutively}
unauditable: the user receives the output of a process she cannot
inspect, hosted on infrastructure she does not control, in a
representation she cannot take elsewhere. Retrieval-based memory over an
owned, structured, inspectable map is the other branch. We do not claim
one branch will win; we claim that for regulated sectors, public
institutions, researchers under ethics oversight, and any person who
regards decades of intellectual life as something to \emph{own}, the
auditable branch is not optional. MOSS demonstrates that it is also
practical. To be fair, synthesis at platform scale has a decisive
advantage MOSS does not contest: zero configuration---the user deploys
and maintains nothing. For the great many who want a memory that simply
works, that convenience wins by default. MOSS addresses the
complementary case, far from marginal in regulated sectors, where
auditability is not negotiable and where the convenience of an
uninspectable, unportable profile is precisely the liability.

\subsection{Sovereignty as an architectural property}

Data sovereignty is usually discussed as policy. MOSS treats it as
architecture: because the memory structure is standard SQL and the
corpus is plain files, the entire system can be relocated---across
clouds, into a national cloud region, onto a laptop---without
transformation. Nothing about the user's memory is
captive to a vendor's embedding space, graph format, or
hosting. In a period when national AI strategies explicitly prioritize
sovereign infrastructure, an agentic memory whose every layer is owned
and portable is directly responsive.

\subsection{Beyond the individual: a companion memory for organizations}

The deployment reported here accompanies one person, but nothing in the
architecture is specific to a single user. The same
machinery---real-time indexing, an inductive ontology, affective and
temporal layers, deterministic auditable retrieval---answers directly to
settings where a human (or a team) must manage complex information in
real time and have an agent both \emph{retrieve} it and \emph{grow} it
as work proceeds: a true companion memory rather than a static archive.
The choice of \emph{what} to index is itself application-dependent: a
legal team might index emotional variation across deposition transcripts
(to study, retrospectively, how a witness's affect
shifts under specific questioning); a clinic or health system might
track the evolution of a case; a high-tempo operations environment such
as an airport might index real-time situational information. The number
and kind of such information layers is unbounded---one can, for
instance, add a layer that flags where quotations or code blocks
occur---and depends entirely on the needs of the person or organization
deploying MOSS. The individual knowledge worker is simply the
\emph{ontologically hardest} case, and solving it first is
methodologically sound: enterprise deployments inherit an ontology (a
law firm has clients and matters; a hospital has patients and episodes),
whereas an individual's corpus mixes research, teaching,
administration, family, finance, and decades of evolving terminology
under no pre-existing schema. The inductive approach forced by the
individual case transfers downward to organizations trivially, while
imposed enterprise ontologies do not transfer upward to lives. It is
worth noting how mainstream assistants have approached the same problem
from the opposite end: they bolt on increasingly heavy ``projects,''
``folders,'' and ``sessions'' that the user must create, name, and maintain
by hand---a manual filing system layered over a stateless model. The
active overlays sketched in Section 6 point the other way: thread
structure is something the system detects and tracks dynamically as work
unfolds, not scaffolding the user is asked to build and tend.

\subsection{An exocortex, on the record}

Cognitive science has long entertained the idea that memory artifacts
can be constitutive parts of a cognitive system rather than mere tools
[Clark \& Chalmers, 1998]. A continuously indexed, agent-navigable
record of one person's working life---consulted daily,
trusted, and inspectable---is arguably the most concrete instantiation
of that idea yet reported: an \emph{exocortex} in the literal sense of
an external, organized, queryable memory substrate coupled to ongoing
thought. We flag, without developing here, that MOSS therefore
constitutes an empirical site for extended-mind research, not only an
engineering artifact; a companion paper takes up that thread.

\subsection{Limitations}

Beyond the evidentiary limits of Section 4.4: (i) determinism in MOSS is
a property of retrieval \emph{execution}, not of the whole
pipeline---query formulation remains agentic and therefore stochastic;
we consider this confinement of non-determinism a feature, but it must
not be overread. (ii) The inductive ontology was built for one corpus;
its construction cost for new corpora, while increasingly automatable,
is not yet characterized. (iii) Multimodal enrichment (audio, video,
image) is in development; the present system is text-complete
(image-only PDFs currently being transcribed) but not media-complete.
(iv) The single-user deployment leaves multi-tenant isolation and
memory-poisoning defenses [cf. Bhardwaj, 2026] as future
engineering.

\section{Ongoing Development: The \emph{Métacalque} Principle}

The overlays of Section 3.3 are realized, in the production system, as
columns and tables of the main database. In parallel, we are developing
a unified \emph{overlay layer}---a separate analytical store in which
every kind of annotation is expressed as one type of object---governed
by a single principle we name here in order to define it precisely: the
\emph{métacalque}. We give it a dedicated section both because it guides
the system's near-term development and because, to our
knowledge, it has no precedent in computational form---a claim we make
carefully and qualify below.

\subsection{A cartographic intuition and its disciplinary lineages}

The \emph{métacalque} did not originate in any of the traditions we are about to invoke. It began in the cartographic relation that runs through this paper---\emph{the database is the map; the documents are the territory}---and in the practice of laying transparent, removable, combinable sheets over that map. Only afterward did we recognize that the same structure has a long \emph{descriptive} life in literary theory and a mature \emph{operational} life in geomatics. We present both not as sources from which the principle is derived, but as independent confirmations that it names something general.

\textbf{A descriptive lineage in literary and semiotic theory.} The idea that one text can be read through multiple superimposed strata of meaning is old and well developed---but, until now, exclusively \emph{descriptive}. Greimas's \emph{isotopies} [Greimas, 1966] posit recurrent semantic layers a reader follows through a text; Eco took up the notion in his work on interpretation [Eco, 1979]; Riffaterre's semiotics of reading [Riffaterre, 1978] proceeds through distinct levels; functional-linguistic stratification, after Halliday [Halliday, 1994], models language itself as superimposed strata. Most paradigmatically, Genette's \emph{Palimpsests} [Genette, 1982/1997] treats a text as a layered surface: his categories of transtextuality---intertextuality, paratextuality, metatextuality, architextuality, hypertextuality---are, in effect, distinct reading grids applied to the same text, each bringing a different stratum of relation into view. Reading \emph{under} the lens of hypertextuality surfaces what a text derives from; reading it under intertextuality surfaces the citations co-present within it. The palimpsest---a parchment rewritten while older writing remains visible beneath---is precisely the figure of a text legible through stacked, semi-transparent layers.\footnote{This conception belongs to the study of texts, well outside computer science, and informs the present work indirectly rather than by importation. Genette himself noted that the relation between his sense of ``hypertext'' and the meaning the term later acquired in electronic textuality would deserve a separate study---a bridge he named but did not cross. For an application of Genette's transtextual framework to another medium---music---see [Lacasse, 2018], which develops a model of \emph{transphonography} in a volume itself devoted to the music palimpsest.}

\textbf{An operational lineage in geomatics.} Where the literary tradition only describes layered reading, one discipline has practiced it \emph{operationally} for decades: geomatics---the engineering science of measuring and analysing geographically referenced data, as distinct from geography proper, which studies human relations to the earth.\footnote{The term \emph{géomatique} is itself a Québécois coinage, popularized by the surveyor Michel Paradis in 1982; the world's first university program in geomatics was founded at Université Laval in 1986.} Multicriteria spatial analysis lays superimposed \emph{thematic layers} over a common coordinate substrate, selects within a layer by attribute, and combines several layers by \emph{weighted overlay} (map algebra) into a \emph{synthesis} or \emph{suitability} map [Malczewski, 1999]. The vocabulary maps onto ours almost term for term---layer/\emph{calque}, select-by-attribute/\emph{filter}, weighted-overlay/\emph{composite}, suitability-map/synthesis map---and Bertin's \emph{Semiology of Graphics} [Bertin, 1967/1983] supplies the underlying grammar: every map is an implantation of points, lines, or areas, exactly the forms a retrieval surface can take. Crucially, this layering is not a reading a human performs after the fact; it is an executable computation that returns a result. Geomatics is thus the operational precedent the literary tradition is not.\footnote{The same structure---a common coordinate axis bearing superimposed, independently queryable annotation layers---recurs beyond geomatics. Genome browsers organize annotations as \emph{tracks} aligned to a reference sequence and backed by a queryable relational store [Kent et al., 2002]; multi-wavelength astronomy overlays images of the same sky region across spectral bands. We take this recurrence as evidence that the \emph{métacalque} isolates a general pattern rather than a local device.}

\subsection{From description to control: the operational turn}

These two lineages leave a precise gap between them. The literary tradition \emph{describes} how a human reader construes a text through layers, but it does not act---it situates and classifies. Geomatics \emph{acts}---its overlays are executable computations---but over physical space, on geographically referenced measurements, never over text or memory. Neither applies operational, layered analysis to a \emph{textual, conversational, and documentary} corpus treated as the memory of a person or an organization.

The \emph{métacalque} occupies exactly that gap. It inherits the \emph{textual} object of the literary tradition and the \emph{operational} character of the geomatic one, and fuses them: an overlay is no longer a lens a human applies to interpret a text, nor a thematic layer over terrain, but an executable stratum under which an agent navigates and retrieves over a corpus. The shift is from hermeneutics to mechanism: an overlay does not merely let a reader see a stratum of meaning; it governs what an agent retrieves, in what order, under what weighting. This is the same gesture the paper makes in reclaiming ``semantic'' from distributional similarity for structured meaning (Section 2.5): a notion long treated as soft and interpretive is given explicit, structured, executable form. To our knowledge, no prior work in artificial intelligence or information retrieval has cast multi-layered reading as an operational primitive of agent memory.

\subsection{An operational definition}

A corpus under the \emph{métacalque} principle admits arbitrarily many
\emph{overlays} (Fr. \emph{calques}), each an executable stratum with
these properties:

\begin{enumerate}
\def\labelenumi{\arabic{enumi}.}
\item
  \textbf{Interval-bounded extent.} An overlay's
  units---its \emph{extents} (Fr. \emph{nappes})---are defined by
  intervals over an explicit axis (token or timestamp ranges), not by a
  fixed, pre-imposed segmentation. The boundaries are properties of the
  overlay, not of the text.
\item
  \textbf{Superimposition and overlap.} Overlays coexist over the same
  corpus and may overlap freely; a given point belongs to as many
  overlays as are relevant, none subordinate to another.
\item
  \textbf{Discrete or continuous boundaries.} Some overlays are discrete
  (a code block either is or is not present); others are continuous,
  where one extent fades into the next (affect shifts gradually, so its
  boundaries are soft and admit a measured overlap rather than a hard
  token cut). Each extent may further carry an internal variation
  profile---an \emph{envelope} (Fr. \emph{enveloppe}), borrowing the
  term from sound synthesis---recording how a value rises and falls
  across the extent.
\item
  \textbf{Descriptive-to-operational status.} An overlay is not an
  annotation to be read but a control structure to be executed: it
  participates in the formulation and weighting of retrieval (Section
  3.4), making ``reading under a lens'' an action the agent performs
  rather than a description after the fact.
\end{enumerate}

The vocabulary is layered to match: the \emph{métacalque} is the
principle; a \emph{calque} (overlay) is one stratum; a \emph{nappe}
(extent) is one bounded occurrence within it; an \emph{enveloppe}
(envelope) is the internal profile of an extent.

\subsection{Maturity: what works, and what is being designed}

The principle is being brought into being at two distinct stages of
maturity, and we are explicit about which is which.

\emph{Implemented and validated, not yet integrated.} In a separate
analytical store, several overlays are already built, populated, and
queryable in SQL, though not yet wired into the production retrieval
pipeline or the agent's operating instructions: an
\emph{affective} overlay (continuous, on Russell's
valence--activation plane, with envelopes), a \emph{code} overlay, a
\emph{repérage} (structural-anchor) overlay derived from document
outlines, a \emph{concepts} overlay, a \emph{genre\_document} overlay,
and a \emph{concept\_cluster} overlay obtained by geometric clustering
of segment embeddings. The structural-anchor overlay is, conceptually,
the one the others lean on: it lays down the coordinate system---the
token and section offsets---to which every other overlay refers. An
\emph{insertion} overlay (marking externally injected documents) is
defined and awaiting population.

\emph{Under active design.} Other overlays are genuinely in conception.
The most ambitious is a \emph{projet} (project) overlay with
\emph{implicit, dynamic} boundaries: as a conversation turns to a
project, an opening marker is laid on the project overlay; when the
discussion closes---or is interrupted by an unrelated \emph{idée}
(idea), which opens its own overlay on top---a closing marker is laid,
and a later return reopens the same project. Retrieval then becomes a
matter of moving cleanly between projects and ideas, suspending and
resuming each, because the agent knows where each begins and ends.
Realizing such boundaries---often inferred from context rather than read
off a fixed token---together with the reconciliation of overlapping
overlays, is the substance of ongoing work.

\subsection{Status and priority}

We name the \emph{métacalque} here, with a dated definition, precisely
because it spans the boundary between the working and the planned: part
is implemented and queryable, part is still being designed. We state the
principle in the open both to invite scrutiny and to establish, by
publication, that the transposition of layered reading from a
descriptive device of literary theory into an operational primitive of
agent memory is, as of this writing, without precedent in the
computational literature. A full treatment---the determinate
construction of certain overlays, the handling of soft boundaries, the
reconciliation of overlapping extents, and their integration into query
formulation---is the subject of a dedicated paper in preparation.

\section{Future Work}

Three lines are underway. \textbf{Benchmarking:} controlled A/B
evaluation of MOSS against vector-RAG baselines and published systems,
measuring accuracy, tokens, latency, and energy---matched capacity by
capacity to appropriate benchmark types (note~\ref{fn:bench}). \textbf{Multi-instance
deployment:} a multi-researcher academic deployment (one sovereign
instance per researcher) under research-ethics oversight, extending the
architecture from one cognitive life to several and producing the first
comparative data on personalization of the inductive
ontology.\footnote{This deployment is supported by a grant from OBVIA (Observatoire international sur les impacts sociétaux de l'intelligence artificielle et du numérique). The funded project, beginning 1 July 2026, brings together researchers from different disciplines at Université Laval---including a specialist in the impact of AI in the workplace---to put the system, developed over the preceding eighteen months, to work as a research instrument. It is MOSS's first funded research application.} \textbf{Multimodal completion:} extending
enrichment to the audio, video, and image holdings already catalogued.
Three companion papers are in preparation: an epistemic framework of
memory (the axioms, properties, and operations underlying Section 3.1),
an extended-mind analysis of the human--agent--memory system (Section
5.4), and a full treatment of the \emph{métacalque} overlay layer
(Section 6). \textbf{Temporal structuring.} Underlying all of this is
the parameter we have, in a sense, been describing throughout:
\emph{time}. It is worth saying plainly, as a closing thought, what the
preceding sections imply at every turn---that a store one merely queries
is not yet a memory. What makes a memory \emph{is} time: order,
validity, succession, and the revisability of what was once held true.
Seen this way, the system already keeps not one time but several. The
content it holds carries the time at which something was \emph{said},
the time at which it was first \emph{produced}, and the time at which it
is \emph{true of the world}---at least three clocks that collapse into
one for a fresh remark and separate for a quotation. To these the memory
adds times of its own: when a fact came to be \emph{known} or revised,
and when it was \emph{indexed} and so made findable. A retrieval system
needs none of this; a memory cannot do without it. Giving these times a
single, principled structure---so that the system reasons over
\emph{when}, not merely over \emph{what}---is the organizing work now
underway, and the one we take to matter most.

\section{Conclusion}

We have presented MOSS, an agentic memory architecture that is auditable
by construction, agnostic on every axis that produces lock-in, and
validated not by a benchmark but by about a year of continuous
production as the working memory of a scholar's life.
Its design inverts the assumptions of mainstream RAG: structure over
similarity, query-time sovereignty over storage-time weighting, an
inspectable map over an opaque synthesis, and a corpus the user owns
over a profile the vendor hosts. Read against the longer history, MOSS
is the agentic heir of the memex lineage---the database of a life,
finally given its navigator. As AI agents move from sessions to years,
we believe these inversions will come to look less like a contrarian
position and more like a precondition.

\section*{Acknowledgments} The first author was partially funded by \href{https://www.obvia.ca/en}{OBVIA} for the AIter Ego project described in Section 7. We would like to thank Alexandre Sasseville and the \href{https://www.ulaval.ca/en/research/office-of-the-vice-rector-research-and-innovation/community-liaison-office}{BLUM} team, as well as the \href{https://www.oicrm.ulaval.ca/}{OICRM} and \href{https://web.iid.ulaval.ca/}{IID} at Université Laval for their support.

\section*{References}

\refentry{Agand, P. (2026). Neuro-symbolic financial reasoning via deterministic fact ledgers and adversarial low-latency hallucination detector. arXiv:2603.04663.}
\refentry{Archish, S., Agarwal, M., Garg, A., Kayal, N., \& Shiragur, K. (2026). On strengths and limitations of single-vector embeddings. arXiv:2603.29519.}
\refentry{Barnett, S., Kurniawan, S., Thudumu, S., Brannelly, Z., \& Abdelrazek, M. (2024). Seven failure points when engineering a retrieval augmented generation system. \emph{CAIN 2024}, 194--199. \url{https://doi.org/10.48550/arXiv.2401.05856}}
\refentry{Bender, E. M., \& Koller, A. (2020). Climbing towards NLU: On meaning, form, and understanding in the age of data. \emph{ACL 2020}. \url{https://doi.org/10.18653/v1/2020.acl-main.463}}
\refentry{Bertin, J. (1983). \emph{Semiology of graphics: Diagrams, networks, maps} (W. J. Berg, Trans.). University of Wisconsin Press. (Original work published 1967.)}
\refentry{Bhardwaj, V. P. (2026). SuperLocalMemory: Privacy-preserving multi-agent memory with Bayesian trust defense against memory poisoning. arXiv:2603.02240.}
\refentry{Biswal, A., Lei, C., Qin, X., Li, A., Narayanaswamy, B.M., \& Kraska, T. (2026). AgentSM: Semantic memory for agentic text-to-SQL.~arXiv:2601.15709.}
\refentry{Braun, V., \& Clarke, V. (2006). Using thematic analysis in psychology. \emph{Qualitative Research in Psychology}, 3(2), 77--101. \url{https://doi.org/10.1191/1478088706qp063oa}}
\refentry{Bush, V. (1945). As we may think. \emph{The Atlantic Monthly}, July 1945.}
\refentry{Chhikara, P., Khant, D., Aryan, S., Singh, T., \& Yadav, D. (2025). Mem0: Building production-ready AI agents with scalable long-term memory. arXiv:2504.19413.}
\refentry{Clark, A., \& Chalmers, D. (1998). The extended mind. \emph{Analysis}, 58(1), 7--19. \url{https://doi.org/10.1093/analys/58.1.7}}
\refentry{Du, P. (2026). Memory for autonomous LLM agents: Mechanisms, evaluation, and emerging frontiers.~arXiv:2603.07670.}
\refentry{Eco, U. (1979). \emph{The role of the reader: Explorations in the semiotics of texts}. Indiana University Press.}
\refentry{Gao, Y., Xiong, Y., Gao, X., Jia, K., Pan, J., Bi, Y., Dai, Y., Sun, J., Guo, Q., Wang, M., \& Wang, H. (2023). Retrieval-augmented generation for large language models: A survey. arXiv:2312.10997.}
\refentry{Gemmell, J., Bell, G., Lueder, R., Drucker, S., \& Wong, C. (2002). MyLifeBits: Fulfilling the memex vision. \emph{ACM Multimedia 2002}. \url{https://doi.org/10.1145/641007.641053}}
\refentry{Gemmell, J., Bell, G., \& Lueder, R. (2006). MyLifeBits: A personal database for everything. \emph{Communications of the ACM}, 49(1), 88--95. \url{https://doi.org/10.1145/1107458.1107460}}
\refentry{Genette, G. (1982/1997). \emph{Palimpsests: Literature in the second degree} (C. Newman \& C. Doubinsky, Trans.). University of Nebraska Press. (Original work \emph{Palimpsestes}, Seuil, 1982.)}
\refentry{GibsonAI (2025). \emph{Memori: An open-source SQL-native memory engine for LLMs and AI agents} [Computer software]. \url{https://github.com/GibsonAI/memori}}
\refentry{Greimas, A. J. (1966). \emph{Sémantique structurale: Recherche de méthode}. Larousse.}
\refentry{Gurrin, C., Smeaton, A. F., \& Doherty, A. R. (2014). LifeLogging: Personal big data. \emph{Foundations and Trends in Information Retrieval}, 8(1), 1--125. \url{http://dx.doi.org/10.1561/1500000033}}
\refentry{Halliday, M. A. K. (1994). \emph{An introduction to functional grammar} (2nd ed.). Edward Arnold.}
\refentry{Han, J., Cheung, A., Wei, Y., Yu, Z., Wang, X., Zhu, B., \& Yang, Y. (2025). RAG meets temporal graphs: Time-sensitive modeling and retrieval for evolving knowledge.~arXiv:2510.13590.}
\refentry{Hu, Y., Liu, S., Yue, Y., Zhang, G., Liu, B., Zhu, F., Lin, J., Guo, H., Dou, S., Xi, Z., Jin, S., Tan, J., Yin, Y., Liu, J., Zhang, Z., Sun, Z., Zhu, Y., Sun, H., Peng, B., Cheng, Z., Fan, X., Guo, J., Yu, X., Zhou, Z., Hu, Z., Huo, J., Wang, J., Niu, Y., Wang, Y., Yin, Z., Hu, X., Liao, Y., Li, Q., Wang, K., Zhou, W., Liu, Y., Cheng, D., Zhang, Q., Gui, T., Pan, S., Zhang, Y., Torr, P., Dou, Z., Wen, J., Huang, X., Jiang, Y., \& Yan, S. (2025). Memory in the age of AI agents.~arXiv:2512.13564.}
\refentry{Hu, Z., Zhu, Q., Yan, H., He, Y., \& Gui, L. (2026). Beyond RAG for agent memory: Retrieval by decoupling and aggregation.~arXiv:2602.02007.}
\refentry{Huyen, C. (2024). \emph{AI engineering: Building applications with foundation models}. O'Reilly. (pp. 258--261 on lexical vs. semantic retrieval.)}
\refentry{Jiang, D., Li, Y., Wei, S., Yang, J., Kishore, A., Zhao, A., Kang, D., Hu, X., Chen, F., Li, Q., \& Li, B. (2026). Anatomy of agentic memory: Taxonomy and empirical analysis of evaluation and system limitations.~arXiv:2602.19320.}
\refentry{Kent, W. J., Sugnet, C. W., Furey, T. S., Roskin, K. M., Pringle, T. H., Zahler, A. M., \& Haussler, D. (2002). The human genome browser at UCSC.~\emph{Genome Research, 12}, 996--1006. \url{https://doi.org/10.1101/gr.229102}}
\refentry{Killingback, J., Rafiee, M., Manas, M. C., \& Zamani, H. (2026). Scaling laws for embedding dimension in information retrieval.~arXiv:2602.05062.}
\refentry{Lacasse, S. (2018). Toward a model of transphonography. In L. Burns \& S. Lacasse (Eds.), \emph{The pop palimpsest: Intertextuality in recorded popular music} (pp. 9--60). University of Michigan Press.}
\refentry{Lenci, A. (2018). Distributional models of word meaning. \emph{Annual Review of Linguistics}, 4, 151--171. \url{https://doi.org/10.1146/annurev-linguistics-030514-125254}}
\refentry{Lin, T., Luo, Y., \& Tang, N. (2026). AnnoRetrieve: Efficient structured retrieval for unstructured document analysis.~arXiv:2604.02690.}
\refentry{Liu, N. F., Lin, K., Hewitt, J., Paranjape, A., Bevilacqua, M., Petroni, F., \& Liang, P. (2024). Lost in the middle: How language models use long contexts.~\emph{Transactions of the Association for Computational Linguistics, 12}, 157--173. \url{https://doi.org/10.1162/tacl_a_00638}}
\refentry{Luo, H., Haihong, E., Chen, G., Lin, Q., Guo, Y., Xu, F., Kuang, Z., Song, M., Wu, X., Zhu, Y., \& Luu, A. (2025). Graph-R1: Towards agentic GraphRAG framework via end-to-end reinforcement learning.~arXiv:2507.21892.}
\refentry{Maharana, A., Lee, D.-H., Tulyakov, S., Bansal, M., Barbieri, F., \& Fang, Y. (2024). Evaluating very long-term conversational memory of LLM agents. arXiv:2402.17753.}
\refentry{Malczewski, J. (1999). \emph{GIS and multicriteria decision analysis}. John Wiley \& Sons.}
\refentry{Mattek, A. M., Wolford, G. L., \& Whalen, P. J. (2017). A mathematical model captures the structure of subjective affect. \emph{Perspectives on Psychological Science}, 12(3), 508--526. \url{https://doi.org/10.1177/1745691616685863}}
\refentry{Menschikov, M., Evseev, D., Dochkina, V., Kostoev, R., Perepechkin, I., Anokhin, P., Semenov, N., \& Burnaev, E. (2026). PersonalAI: A systematic comparison of knowledge graph storage and retrieval approaches for personalized LLM agents. \emph{IEEE Access}, \emph{14}, 58262--58281. \url{https://doi.org/10.1109/ACCESS.2026.3682941}}
\refentry{OpenAI (2026, June 4). Dreaming: Better memory for a more helpful ChatGPT. \url{https://openai.com/index/chatgpt-memory-dreaming/}}
\refentry{Packer, C., Fang, V., Patil, S. G., Lin, K., Wooders, S., \& Gonzalez, J. (2023). MemGPT: Towards LLMs as operating systems.~arXiv:2310.08560.}
\refentry{Pan, J. J., Wang, J., \& Li, G. (2023). Survey of vector database management systems.~\emph{The VLDB Journal, 33}, 1591--1615. \url{https://doi.org/10.48550/arXiv.2310.14021}}
\refentry{Peng, B., Zhu, Y., Liu, Y., Bo, X., Shi, H., Hong, C., Zhang, Y., \& Tang, S. (2024). Graph retrieval-augmented generation: A survey.~\emph{ACM Transactions on Information Systems, 44}, 1--52. \url{https://doi.org/10.48550/arXiv.2408.08921}}
\refentry{Quillian, M. R. (1968). Semantic memory. In M. Minsky (Ed.), \emph{Semantic information processing} (pp. 227--270). MIT Press.}
\refentry{Rasmussen, P., Paliychuk, P., Beauvais, T., Ryan, J., \& Chalef, D. (2025). Zep: A temporal knowledge graph architecture for agent memory. arXiv:2501.13956.}
\refentry{Riffaterre, M. (1978). \emph{Semiotics of poetry}. Indiana University Press.}
\refentry{Russell, J. A. (1980). A circumplex model of affect. \emph{Journal of Personality and Social Psychology}, 39(6), 1161--1178. \url{https://psycnet.apa.org/doi/10.1037/h0077714}}
\refentry{Sahlgren, M. (2008). The distributional hypothesis. \emph{Italian Journal of Linguistics}, 20(1), 33--53.}
\refentry{Sanmartin, D. (2024). KG-RAG: Bridging the gap between knowledge and creativity.~arXiv:2405.12035.}
\refentry{Tulving, E. (1972). Episodic and semantic memory. In E. Tulving \& W. Donaldson (Eds.), \emph{Organization of memory}. Academic Press.}
\refentry{Weller, O., Boratko, M., Naim, I., \& Lee, J. (2025). On the theoretical limitations of embedding-based retrieval. arXiv:2508.21038.}
\refentry{Xu, W., Liang, Z., Mei, K., Gao, H., Tan, J., \& Zhang, Y. (2025). A-MEM: Agentic memory for LLM agents.~arXiv:2502.12110.}
\refentry{Yang, C., Zhou, C., Xiao, Y., Dong, S., Zhuang, L., Zhang, Y., Wang, Z., Hong, Z., Yuan, Z., Xiang, Z., Chen, S., Zhou, H., Zhang, Q., Liu, N., Su, J., Wang, X., Chang, Y., \& Huang, X. (2026). Graph-based agent memory: Taxonomy, techniques, and applications.~arXiv:2602.05665.}
\refentry{Yu, Y., Yao, L., Xie, Y., Tan, Q. S., Feng, J., Li, Y., \& Wu, L. (2026). Agentic memory: Learning unified long-term and short-term memory management for large language model agents.~arXiv:2601.01885.}
\refentry{Zhang, Q., Chen, S., Bei, Y., Yuan, Z., Zhou, H., Hong, Z., Dong, J., Chen, H., Chang, Y., \& Huang, X. (2025). A survey of graph retrieval-augmented generation for customized large language models.~arXiv:2501.13958.}
\refentry{Zhang, X., Wang, D., Xu, K., Zhu, Q., \& Che, W. (2026). Scaling laws for agent harnesses via effective feedback compute. arXiv:2605.29682.}
\refentry{Zhang, Z., Dai, Q., Bo, X., Ma, C., Li, R., Chen, X., Zhu, J., Dong, Z., \& Wen, J. (2024). A survey on the memory mechanism of large language model-based agents.~\emph{ACM Transactions on Information Systems, 43}, 1--47. \url{https://doi.org/10.48550/arXiv.2404.13501}}
\refentry{Zhu, Z., Huang, T., Wang, K., Ye, J., Chen, X., \& Luo, S. (2025). Graph-based approaches and functionalities in retrieval-augmented generation: A comprehensive survey.~\emph{ACM Computing Surveys}. \url{https://doi.org/10.48550/arXiv.2504.10499}}

\end{document}